\pdfoutput=1
\PassOptionsToPackage{table,xcdraw}{xcolor}
\documentclass[sigconf,nonacm,balance=false]{acmart}

\usepackage{graphics}
\usepackage{graphicx}
\usepackage{xspace}
\usepackage{subcaption}
\usepackage{amsmath}
\usepackage{amsthm}
\usepackage{mathrsfs}
\usepackage{array}
\usepackage{textcomp}
\usepackage{weiwAlgorithm}
\usepackage{diagbox}
\usepackage{booktabs}
\usepackage{listings}
\usepackage{tcolorbox}
\usepackage{setspace}
\usepackage{multirow}
\usepackage{tabularx}
\usepackage{makecell}
\usepackage{enumitem}
\usepackage{float}
\usepackage{placeins}
\usepackage{stfloats}
\usepackage{afterpage}
\usepackage{wrapfig}
\usepackage{pgfplots}
\usepackage[normalem]{ulem}

\pgfplotsset{compat=1.17}
\useunder{\uline}{\ul}{}
\makeatletter
\patchcmd{\maketitle}
  {\@specialsection{\keywordsname}}
  {\vspace*{-2mm}\@specialsection{\keywordsname}}
  {}{\PackageError{ctrlg}{Cannot adjust spacing above Keywords}{Check the acmart top-matter definition.}}
\patchcmd{\maketitle}
  {\noindent\@keywords\par}
  {\noindent\@keywords\par\vspace*{-2mm}}
  {}{\PackageError{ctrlg}{Cannot adjust spacing below Keywords}{Check the acmart top-matter definition.}}
\makeatother

\usepackage{amsmath,amsfonts,bm}

\def\1{\bm{1}}

\DeclareMathAlphabet{\mathsfit}{\encodingdefault}{\sfdefault}{m}{sl}
\SetMathAlphabet{\mathsfit}{bold}{\encodingdefault}{\sfdefault}{bx}{n}

\definecolor{kgblue}{RGB}{55,115,179}
\definecolor{wikigreen}{RGB}{46,153,83}
\definecolor{weborange}{RGB}{220,130,30}

\newtheorem{definition}{Definition}

\newcommand{\prag}{Ctrl-G\xspace}
\newcommand{\myparagraph}[1]{\noindent \textbf{#1}.}
\newcommand{\myparagraphunderline}[1]{\noindent \underline{#1.}}

\AtBeginDocument{%
  }

\setcopyright{none}
\renewcommand\footnotetextcopyrightpermission[1]{}
\title{Ctrl-G: Controlling Structural and Interaction Exposure in Knowledge Graph Augmented LLM Reasoning}

\author{Xingyu Tan}
\affiliation{\institution{UNSW \& CSIRO}\city{Sydney}\country{Australia}}
\email{xingyu.tan@unsw.edu.au}
\author{Xiaoyang Wang}
\affiliation{\institution{UNSW}\city{Sydney}\country{Australia}}
\email{xiaoyang.wang1@unsw.edu.au}
\author{Qing Liu}
\affiliation{\institution{CSIRO}\city{Hobart}\country{Australia}}
\email{q.liu@csiro.au}
\author{Xiwei Xu}
\affiliation{\institution{CSIRO}\city{Sydney}\country{Australia}}
\email{xiwei.xu@csiro.au}
\author{Xin Yuan}
\affiliation{\institution{CSIRO \& UNSW}\city{Sydney}\country{Australia}}
\email{xin.yuan@csiro.au}
\author{Liming Zhu}
\affiliation{\institution{CSIRO}\city{Sydney}\country{Australia}}
\email{liming.zhu@csiro.au}
\author{Wenjie Zhang}
\affiliation{\institution{UNSW}\city{Sydney}\country{Australia}}
\email{wenjie.zhang@unsw.edu.au}
\renewcommand{\shortauthors}{Tan et al.}
\hypersetup{pdfauthor={Xingyu Tan, Xiaoyang Wang, Qing Liu, Xiwei Xu, Xin Yuan, Liming Zhu, Wenjie Zhang}}

\begin{document}

\begin{abstract}
Knowledge graphs (KGs) provide structured evidence for Web question answering with large language models (LLMs), but sending retrieved evidence to a remote LLM can expose sensitive graph information. Existing privacy treatments use entity anonymization and relation-centric abstraction to preserve reasoning cues, but transmitted topology remains visible and disclosure accumulates across repeated calls. Distinctive topology can support re-identification of masked entities (structural exposure), while successive calls accumulate additional paths and exploration traces into a broader view of the KG (interaction exposure). Simply sending less evidence is insufficient because restricted, stepwise views may omit the bridge paths required for multi-hop reasoning. To address this, we propose \textbf{Ctrl-G}, to the best of our knowledge the first KG-RAG framework to jointly control \textbf{structural and interaction exposure} to a remote LLM while supporting accurate multi-hop reasoning. To control structural exposure, Ctrl-G locally constructs query-conditioned sanitized views through structural aggregation and pruning, limiting the graph evidence transmitted to a remote LLM. A hierarchical controller couples remote planning with local connected-path exploration across topic entities, using question indicators and retrieved experience to prune candidates while grounding and verifying retained evidence against the original KG. To control interaction exposure, local experience memory reuses abstract, verified reasoning traces to guide exploration modes and depths and gate remote LLM calls, reducing repeated disclosure across questions. Ctrl-G achieves the highest accuracy among the compared methods on five of six benchmarks, exceeding the strongest baseline by up to 13.8 points across model configurations. Compared with semantic masking on the same CWQ questions, Ctrl-G reduces topology-based re-identification success from 41.3\% to 12.8\% with a 0.7-point decrease in accuracy. It also averages 6.5 remote calls per question on CWQ, 70\% fewer than ToG's 21.7.
\end{abstract}


\raggedbottom
\maketitle
\afterpage{\flushbottom}

\begin{figure}[!t]
  \centering
  \includegraphics[width=0.99\columnwidth]{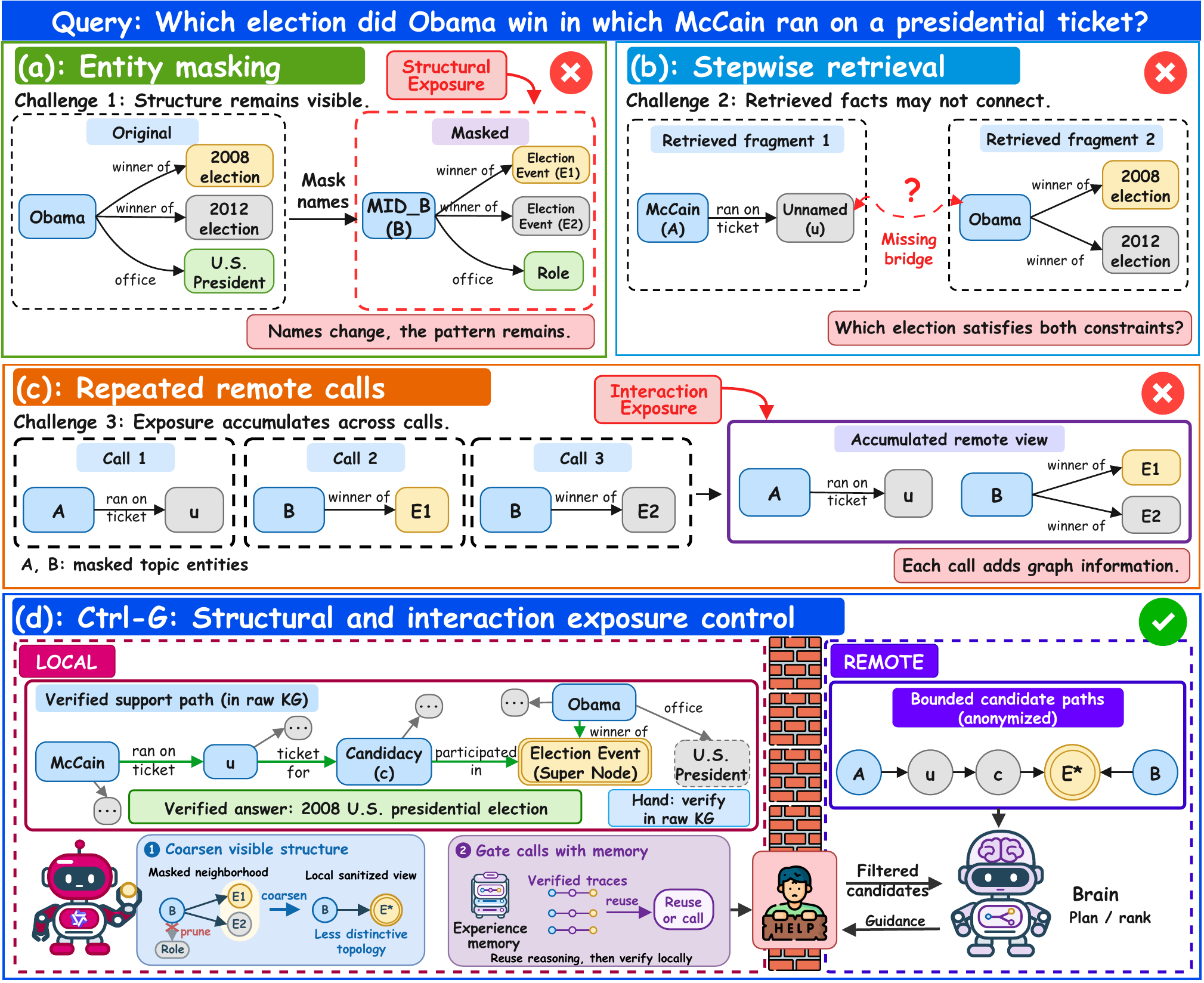}
  \vspace{0.5mm}
  \caption{Overview of existing approaches and Ctrl-G.}
  \vspace{0mm}
  \label{fig:intro_demo}
  \Description{The panels illustrate structural exposure after entity masking, disconnected evidence, cumulative exposure across calls, and Ctrl-G's local Hand and remote Brain.}
\end{figure}

\section{Introduction}
\label{sec:intro}

Large Language Models (LLMs) acquire broad capabilities from large-scale pre-training, but their parametric knowledge is difficult to update and can be incomplete or outdated~\cite{brown2020language,chowdhery2023palm,touvron2023llama}.
Retrieval-Augmented Generation (RAG) addresses this limitation by grounding generation in external evidence at inference time~\cite{baek2023knowledge,gao2023retrieval}.
Knowledge Graphs (KGs) are particularly valuable for Web question answering (QA), where their explicit entities and relations support multi-hop reasoning, evidence tracing, and local verification~\cite{tog1.0sun2023think,rogluo2023reasoning}.
KG-based RAG systems typically link topic entities, retrieve candidate triples or paths, and generate an answer from the retrieved evidence~\cite{tan2025hydra}.
In many deployments, however, the KG is private while reasoning relies on an API-served LLM~\cite{hu2024privacy2}.
Every transmitted triple, candidate path, and search trace then reveals knowledge about the graph to the remote service, which is problematic for enterprise, healthcare, financial, and internal-security KGs~\cite{arog_privacy}.
This raises a central question: how can a system benefit from remote reasoning while limiting what the service learns about the graph and grounding each answer in locally verified KG evidence?

\vspace{1mm}

\myparagraph{Challenges in exposure-controlled KG reasoning}
Private KG-RAG methods combine entity anonymization with relation-centric abstraction to retrieve evidence from transformed KGs~\cite{arog_privacy}.
These methods hide entity identities while preserving relational cues. The connections needed for reasoning can also identify parts of the graph, and repeated calls reveal more of those connections. Figure~\ref{fig:intro_demo}(a--c) illustrates three challenges in controlling this exposure while retaining connected evidence for accurate multi-hop QA.

\vspace{0.3mm}
\myparagraphunderline{Challenge 1: Structural exposure beyond masking}
Masking entity names leaves their connectivity visible. In Figure~\ref{fig:intro_demo}(a), entity $B$ retains links to two election events and an office role. These links can reveal $B$'s identity when matched against an auxiliary graph. Such visible topology, path motifs, and neighborhood statistics constitute \textbf{structural exposure}. Controlling this exposure requires reducing identifying patterns while preserving the relations and cross-entity connections needed for multi-hop reasoning.

\vspace{0.3mm}
\myparagraphunderline{Challenge 2: Reasoning under restricted views}
Restricted views can retain entity-specific evidence while obscuring connections needed for multi-hop reasoning. Independent retrieval and early pruning can miss or discard connecting paths, while structural coarsening can further obscure them. In Figure~\ref{fig:intro_demo}(b), McCain's ticket relation and Obama's election victories are retrieved separately, but neither fragment establishes which election satisfies both constraints. The missing ticket-to-election bridge must still be recovered and verified against the original KG to establish joint support for the answer across the question's topic entities.

\vspace{0.3mm}
\myparagraphunderline{Challenge 3: Repeated interaction exposure}
Limiting evidence per remote call does not control what successive calls reveal together. Question analysis, candidate selection, and recovery from failed searches can expose different fragments of the same graph. In Figure~\ref{fig:intro_demo}(c), these masked fragments gradually form a broader view of its structure, creating \textbf{interaction exposure}. Without reusing exploration experience, similar questions may repeat the same expansions and failed branches, prompting further remote interaction. Controlling this exposure requires deciding when local reasoning or verified experience can replace another remote call, while verifying each answer locally using current evidence from the original KG.

\vspace{1mm}

\myparagraph{Contributions}
To the best of our knowledge, \textbf{Ctrl-G}, shown in Figure~\ref{fig:intro_demo}(d), is the first KG-RAG framework to jointly control structural and interaction exposure to a remote LLM. It controls both what graph information the remote model observes and when further interaction is needed. A remote LLM (Brain) guides reasoning over sanitized views, while a local LLM (Hand) retrieves and verifies evidence against the original KG. Experience memory reuses successful exploration decisions to reduce repeated search and remote calls. Structural sanitization, connected-path reasoning, and memory-guided interaction control address these challenges:

\vspace{0.3mm}
\noindent\raisebox{0.2ex}{\scriptsize$\bullet$}\enspace
\textbf{Structural exposure control.}
To limit structural exposure beyond entity masking, Ctrl-G constructs a local question-specific subgraph $G^{\mathrm{raw}}_Q$ and derives a sanitized view $\tilde{G}_Q$ through structural aggregation and pruning. It coarsens distinctive connectivity and attributes that may support topology-based linkage, transmitting only selected candidate paths to the Brain. This limits exposure while retaining the raw KG for local grounding and verification.

\vspace{0.3mm}

\noindent\raisebox{0.2ex}{\scriptsize$\bullet$}\enspace
\textbf{Dual-LLM hierarchical reasoning.}
To recover connected evidence under restricted views, Ctrl-G coordinates the Brain and Hand through a hierarchical reasoning tree. Remote planning and ranking of sanitized inputs guide local connected-path exploration, with question indicators and retrieved experience informing pruning. The Hand grounds retained paths in the original KG and assesses their joint support for the question. Verification feedback and experience guide further exploration and Brain calls.

\vspace{0.3mm}

\noindent\raisebox{0.2ex}{\scriptsize$\bullet$}\enspace
\textbf{Memory-guided interaction control.}
To control exposure across repeated interactions, Ctrl-G maintains local memory of abstract, verified exploration traces and failure patterns. Retrieved experience guides search modes and depths, avoids failed branches, and enables reuse of question analysis without another Brain call. The controller checks evidence sufficiency before further exploration and records episode outcomes for later questions. This reduces repeated disclosures while preserving local KG verification.

\vspace{0.3mm}

\noindent\raisebox{0.2ex}{\scriptsize$\bullet$}\enspace
\textbf{Adaptability, exposure control, effectiveness, and interaction efficiency.}
\textbf{(a)} Ctrl-G operates as a training-free, plug-and-play framework, with adaptability demonstrated across eight LLM backbones serving as Hand or Brain. It accesses different or updated KGs through retrieval without model retraining.
\textbf{(b)} On matched CWQ questions and backbones, Ctrl-G reduces topology-based re-identification success from 41.3\% under semantic masking to 12.8\%, retaining 64.2\% accuracy versus 64.9\%.
\textbf{(c)} Across model configurations, Ctrl-G achieves the highest reported accuracy on five of six benchmarks among the compared methods, outperforming the strongest baseline by up to 13.8 percentage points on QALD10-en.
\textbf{(d)}~On CWQ, Ctrl-G reduces average remote calls by 70\%, from ToG's 21.7 to 6.5 per question during KG-grounded reasoning.

\begin{figure*}[!t]
    \centering
    \includegraphics[width=0.97\textwidth]{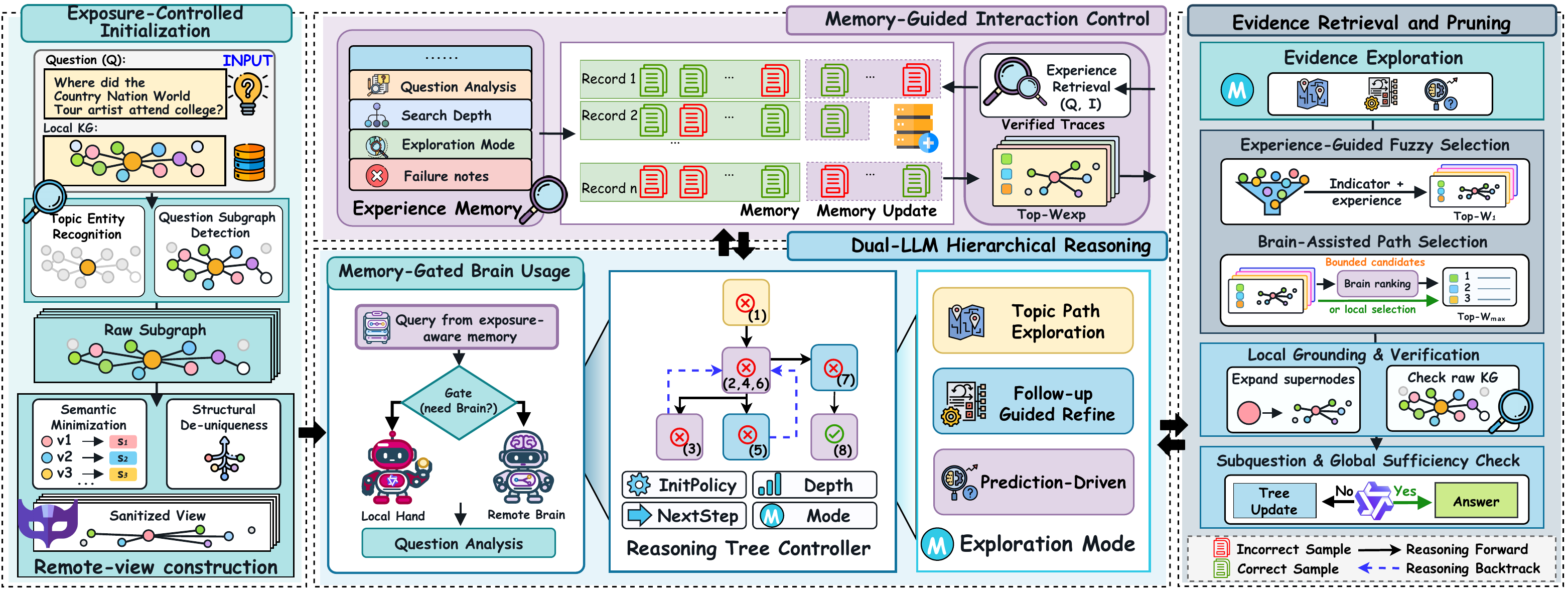}
    \caption{Overview of \prag.
    \textbf{Exposure-controlled initialization} builds raw and sanitized local views. \textbf{Memory-guided interaction control} reuses experience to gate Brain calls. \textbf{Dual-LLM hierarchical reasoning} sets exploration modes and depths. \textbf{Evidence retrieval and pruning} selects connected paths, bounds remote-visible candidates, and verifies paths locally against the raw KG.}
    \vspace{-1mm}
    \label{fig:all}
    \Description{Ctrl-G keeps raw and sanitized graphs local. Memory gates remote planning, and the Hand explores candidates and verifies selected evidence against the raw graph.}
\end{figure*}

\vspace{-2mm}
\section{Related Work}
\label{sec:related_work}

\myparagraph{KG-based RAG}
KGs provide explicit entity relations for grounded and interpretable reasoning~\cite{pan2024unifying}. Early approaches integrate KG knowledge into neural models through pre-training or fine-tuning~\cite{peters2019knowledge}, whereas retrieval-based approaches supply graph evidence at inference time. ToG and ToG-2 use iterative graph exploration guided by LLMs~\cite{tog1.0sun2023think,tog2.0ma2024think}, while StructGPT alternates between reading structured evidence and reasoning over it~\cite{jiang2023structgpt}. Plan-on-Graph (PoG) and DoG organize exploration through planning and candidate evaluation~\cite{plan-on-graph,debated-on-graph}. RoG generates relation-path plans and retrieves matching KG paths as evidence for LLM-based reasoning~\cite{rogluo2023reasoning}. Ctrl-G couples remote reasoning over sanitized candidates with local KG grounding, using verified experience to control subsequent exploration and remote calls across questions.

\myparagraph{Private KG-based RAG}
Private KG-based RAG studies how external LLMs can use graph evidence without receiving raw entity identities. ARoG~\cite{arog_privacy} combines entity anonymization with relation-centric entity abstraction and structured concept paths to guide retrieval over anonymous KGs. These abstractions preserve reasoning cues when entity names are hidden. Ctrl-G extends this focus to transmitted topology and cumulative disclosure across calls through structural sanitization and memory-guided interaction control.

\myparagraph{Memory-augmented reasoning}
Multi-step reasoning methods coordinate intermediate decisions and evidence acquisition through explicit control loops. ReAct interleaves reasoning and actions~\cite{yao2022react}, while Tree of Thoughts explores alternative reasoning states~\cite{yao2024ToT}. Question decomposition supports targeted retrieval by dividing complex questions into simpler subquestions~\cite{pressMeasuringNarrowingCompositionality2023,xuSearchintheChainInteractivelyEnhancing2024}. Memory systems complement these within-question strategies by retaining information across interactions. MemoryBank and MemGPT support long-term interaction histories and external memory management~\cite{memorybank,memgpt}, while Self-Controlled Memory uses a controller to regulate memory access~\cite{wang2023enhancing}. A-Mem, G-Memory, and Evo-Memory further explore memory organization and experience reuse across tasks~\cite{xu2025amem,zhang2025gmemory,wei2025evomemory}. Beyond context retention, Ctrl-G reuses verified experience to guide local search and regulate remote calls and the graph evidence exposed during successive reasoning steps.

\vspace{-2mm}
\section{Preliminaries}\label{sec:prelim}
A KG $G(E,R)$ consists of entities $E$ and relations $R$. Its facts are triples $G(E,R)=\{(e_h,r,e_t)\mid e_h,e_t\in E, r\in R\}$, where $e_h$ and $e_t$ are the head and tail entities and $r$ specifies the connecting relation.

\begin{definition}[Reasoning Path] 
Given a KG $G$,
    a reasoning path within $G$ is defined as a connected sequence of knowledge triples, represented as: ${\text{Path}}_{G}(e_1, e_{l+1}) = \{(e_1,r_1,e_2),(e_2,r_2,e_3),\ldots,(e_l,r_l,e_{l+1})\}$, where $l$ denotes the path length, i.e., ${\rm{length}}({\text{Path}}_{G}(e_1, e_{l+1})) = l$, counting the KG edges traversed between the two endpoint entities.

\end{definition}

\begin{definition}[Entity Path]
Given a KG $G$ and an entity list $\text{list}_e$ = [$e_1, e_2, e_3, \ldots, e_l$], the entity path of $list_e$ is defined as a connected sequence of reasoning paths, which is denoted as 
${\text{Path}}_{G}(list_e)$
$= \{{\text{Path}}_{G}(e_1, e_2), $ $
{\text{Path}}_{G}(e_2$, $e_3), \ldots, {\text{Path}}_{G}(e_{l-1}, e_l) \}=\{(e_s, r, e_t)$$| (e_s, r, e_t)$ $ \in {\text{Path}}_{G}(e_{i}, e_{i+1}) $$\land 1 \leq i < l\}$.

\end{definition}

\myparagraph{Structural and interaction exposure}
We characterize remote exposure along two complementary dimensions. \textbf{Structural exposure} concerns the connectivity patterns and attributes observable in transmitted graph evidence. \textbf{Interaction exposure} concerns the accumulation of evidence and reasoning states across repeated remote calls. The former describes what a transmitted view reveals, while the latter captures how successive views expose additional information or enable linkage across observations. Exposure events occur when information is transmitted to the remote model. Local exploration contributes only through transmitted results.

\myparagraph{Exposure-controlled KGQA}
Given a natural language question $Q$ and a locally held KG $G$, the task is to predict an answer $a=f(Q,G)$ supported by reasoning paths in $G$, with assistance from a remote LLM. The objective is accurate KG-grounded answering while controlling the graph content revealed to the remote model and the frequency of remote interaction. Answer grounding and evidence verification remain local, using reasoning paths in the original KG to check the candidate answer and its support against the constraints expressed in the question before final answering.

\vspace{-4mm}
\section{Method}
\label{sec:method}
\vspace{-1mm}

Under this exposure-controlled setting, \prag coordinates a local LLM (Hand) and a remote LLM (Brain) through memory-guided local control.
Both the raw working graph and its sanitized representation remain local.
The Hand controls local exploration and verifies evidence against the raw KG. The Brain supports planning and candidate selection using bounded sanitized inputs.
As shown in Figure~\ref{fig:all}, \prag consists of four modules: exposure-controlled initialization (Section~\ref{sec:method:init}), dual-LLM hierarchical reasoning (Section~\ref{sec:method:routing}), evidence retrieval and pruning (Section~\ref{sec:Method:retrieval_pruning}), and memory-guided interaction control with experience reuse (Section~\ref{sec:memory}).

\vspace{-3mm}
\subsection{Exposure-Controlled Initialization}
\label{sec:method:init}
\vspace{-1mm}
Initialization builds a question-specific raw subgraph for grounding and verification, then derives a sanitized view for retrieval.

\myparagraph{Knowledge grounding}
Given a question $Q$, \prag first identifies a question-centric subgraph $G^{\mathrm{raw}}_Q$ that covers the topic entities in $Q$ and the surrounding KG neighborhood up to $D_{\max}$ hops.

\myparagraphunderline{Topic entity recognition}
To locate question-relevant entities, \mbox{Ctrl-G} uses the local LLM to extract candidate entities from $Q$, then aligns them with KG entities using a dense retrieval model (DRM).
Top-ranked entities form $T(Q)$, which anchors subgraph construction and connected-path exploration in the local KG.

\begin{figure*}
    \centering
    
    \includegraphics[width=0.85\textwidth]{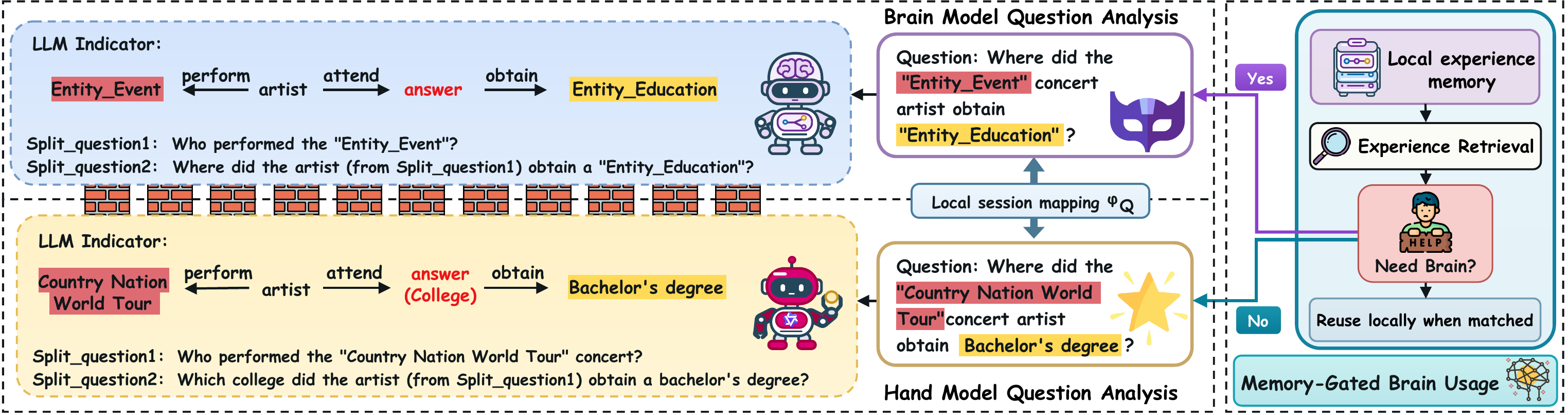}

    \vspace{1mm}
    \caption{Illustration of Memory-Gated Brain and dual-LLM question analysis.}
    \vspace{-2mm}
    \label{fig:question_analysis}
    \Description{Memory-guided question analysis can reuse local experience or use Hand or Brain analysis. A local session mapping connects anonymized and original entities.}
\end{figure*}
\myparagraphunderline{Question subgraph detection}
Given $T(Q)$, \prag constructs the induced subgraph $G^{\mathrm{raw}}_Q \subseteq G$ by bounded expansion around each anchor.
For each $e \in T(Q)$, we collect triples incident to entities within $D_{\max}$ hops of $e$, and then take the union across anchors.

\myparagraph{Remote-view construction}
After obtaining $G^{\mathrm{raw}}_Q$, we construct a sanitized, query-conditioned view $\tilde{G}_Q$ for local exploration and remote-visible candidate selection.
Entity masking alone leaves distinctive topology and rare attribute patterns observable. \prag jointly controls the sanitized view along two axes: \textbf{semantic minimization} limits directly readable content, while \textbf{structural de-uniqueness} coarsens topology and attributes that could link a remote-visible graph region to an auxiliary reference graph.

\myparagraphunderline{Session-local semantic minimization}
We replace entity names with session-specific identifiers, optionally augmented with coarse schema types, and represent relations according to configuration $\pi$.
For each QA session, we sample a fresh secret $s_Q$ and define a temporary mapping $\phi_Q$ (e.g., using a truncated $\mathrm{HMAC}_{s_Q}(e)$ as an entity token). Discarding $s_Q$ and the mapping after the session prevents direct token reuse across sessions. Coarse KG schema types retain weak reasoning constraints without revealing entity names.
Relations are represented by either original labels (utility-oriented mode) or clustered schema labels (stricter exposure-control mode).
We also coarsen sensitive literals: temporal values are discretized to coarse granularity (e.g., $2020$-$03$-$15 \rightarrow 2020$), and numeric values are bucketed into ranges.
These steps limit readable content while preserving coarse KG constraints for multi-hop reasoning.

\myparagraphunderline{Structural sanitization and context minimization}
To reduce structural exposure, \prag sanitizes the anonymized subgraph before constructing remote-visible candidates.
We perform exposure-oriented structure pruning and clustering, building on path-based graph exploration ideas from~\cite{pogtan2025paths}:
we cluster anonymized entities into supernodes and coarsen their relations, then prune entities and relations weakly connected to the topic anchors under a fixed budget.
This coarsens structural features that may support linkage and reduces the candidate structure available for transmission.
The Brain receives selected inputs for planning or ranking: anonymized questions, configured schema or control hints, and candidate paths drawn from $\tilde{G}_Q$ for ranking. Exposure accounting tracks the graph evidence and other inputs transmitted to the Brain throughout successive reasoning steps, for both planning and ranking.

\vspace{-1.5mm}
\subsection{Dual-LLM Hierarchical Reasoning}
\label{sec:method:routing}
Using the initialized graphs, \prag seeks sufficient evidence for multi-hop answering while limiting repeated remote disclosure.
Its controller combines memory-gated question analysis with an experience-guided reasoning tree to coordinate exploration across subquestions.
The Brain $\mathcal{M}_{B}$ analyzes anonymized questions and ranks bounded candidate sets.
The Hand $\mathcal{M}_{H}$ uses $(Q, G^{\mathrm{raw}}_Q)$ and the session mapping $\phi_Q$ to select exploration modes and depths, control state transitions, and verify retained evidence locally.
Algorithm~\ref{algorithm:DualLLMController} in Appendix~\ref{appendix:alg:routing} details the hierarchical reasoning controller.

\myparagraph{Memory-gated Brain usage}
Before invoking $\mathcal{M}_{B}$, the Hand queries the exposure-aware memory in Section~\ref{sec:memory}.
If a confident match is found, it reuses an abstract, locally verified analysis pattern.
Otherwise, the gate chooses between fresh Hand analysis and a Brain call, which receives $(\tilde{Q}, \tilde{T}(Q))$ and bounded memory hints instantiated with current session identifiers.
This gate reduces interaction exposure by avoiding repeated remote analysis of similar questions with reusable decomposition patterns and depth priors.

\myparagraphunderline{Question analysis}
Question analysis decomposes $Q$ into subquestions and produces a skyline indicator $I_Q$ that orders the topic entities and specifies the expected answer position.
The indicator provides a global path sketch that guides subsequent evidence retrieval.
From the indicator, we compute a predicted depth $D_{\text{predict}}$, defined as the maximum distance between the predicted answer and any topic entity.
Figure~\ref{fig:question_analysis} shows an example with $D_{\text{predict}}=2$.
This analysis can be performed by either $\mathcal{M}_{B}$ or $\mathcal{M}_{H}$.
If $\mathcal{M}_{B}$ is invoked, it produces the analysis in the anonymized space, and $\mathcal{M}_{H}$ constructs the de-anonymized counterpart locally via $\phi_Q^{-1}$ for downstream search and verification.
For fresh local analysis, $\mathcal{M}_{H}$ produces the de-anonymized analysis on-device and derives the anonymized counterpart by applying $\phi_Q$ for local retrieval.

\myparagraph{Experience-guided reasoning tree controller}
To coordinate exploration across subquestions, \prag maintains a reasoning tree $\mathcal{T}_Q$.
The node indicators form $\mathcal{I}_Q$, with $I_Q$ at the root. Each node $v$ has an indicator $i_v$ for its subquestion and stores a local state
$
\mathrm{state}(v)=\big(d_v,\; m_v,\; \tilde{P}_v,\; E_v,\; \mathrm{status}(v)\big),
$
where $d_v$ is the depth budget, $m_v \in \{\textsc{Topic},\textsc{Refine},\textsc{Predict}\}$ is the exploration mode, $\tilde{P}_v$ is the anonymized candidate path set on $\tilde{G}_Q$, $E_v$ is the locally verified evidence after de-anonymization and refinement, and $\mathrm{status}(v)$ indicates whether $v$ is active, verified, or pruned. 
The Hand traverses $\mathcal{T}_Q$ top-down and runs a local verification loop at each frontier node: it explores and prunes candidates in the anonymized space, verifies a small retained set locally, and then decides whether to stop or continue at the current frontier under the selected budget.

\myparagraphunderline{InitPolicy and NextStep for mode/depth control}
For each node, the initial exploration configuration $(m_v,d_v)$ is set by an experience-guided \textsc{InitPolicy} before traversal, using retrieved experience (if any) and the depth prior $D_{\text{predict}}$.
If local verification does not succeed, the controller updates $(m_v,d_v)$ via \textsc{NextStep}, deciding whether to increase depth within $D_{\max}$ (or stop early) and whether to keep the current mode or switch modes based on current node-level experience signals and observed local verification failures.

\myparagraphunderline{Exploration, pruning, and verification loop}
At a frontier node $v$, \prag explores an anonymized candidate set $\tilde{P}_v$ on $\tilde{G}_Q$ under the current $(m_v,d_v)$ and, when local selection is insufficient, asks $\mathcal{M}_B$ to rank the bounded candidate set before grounding.
The Hand then \textbf{grounds} the retained candidates in $G^{\mathrm{raw}}_Q$. For unaggregated entities, $\phi_Q^{-1}$ restores the local identifiers. Supernodes are expanded to their original members to recover a consistent connected path. The recovered paths are refined by $\mathcal{M}_H$ into concise evidence $E_v$ for question answering.
The Hand checks $E_v$ against the subquestion's constraints, attaching sufficient evidence to $\mathcal{T}_Q$ or invoking \textsc{NextStep}. Repeated failures and memory warnings trigger branch pruning. It then checks the combined evidence against $Q$ and updates the tree until support is sufficient or the configured search limits prevent further expansion of the current reasoning tree.

\vspace{-1.5mm}
\subsection{Evidence Retrieval and Pruning}
\label{sec:Method:retrieval_pruning}

At each controller step, this module seeks connected paths that jointly support the question's entity and relation constraints.
Retrieval combines evidence exploration and pruning, as detailed in Algorithms~\ref{algorithm:ExploreAndPrune}--\ref{algorithm:EvidencePruning} of Appendix~\ref{appendix:alg:exploration}.
The Hand explores $\tilde{G}_Q$ and applies fuzzy pruning locally to limit remote-visible structure. When Brain-assisted ranking is needed, it transmits the bounded candidate set and grounds the selected paths in the raw KG for verification.

\myparagraph{Evidence exploration}\label{sec:method:evidence_exploration}
To combine KG-grounded search with the LLM's reasoning and parametric knowledge, \prag uses three complementary modes: Topic-path connects topic entities, Follow-up guided refine generates targeted questions for missing evidence, and Prediction-driven proposes answer or bridge entities as new search anchors.
After each mode, candidates are pruned and verified against the raw KG.
The controller stops at sufficient support or adjusts depth or mode using verification feedback and experience.

\myparagraphunderline{Topic-path exploration}
Given $\tilde{G}_Q$, the anchor set $T(Q)$, and the skyline indicator $I_{Q}$, \prag begins at depth $D = \min(D_{\text{predict}}, D_{\max})$ and identifies candidate paths that cover all topic entities in the indicated order.
We use tree-structured bidirectional breadth-first search (BiBFS) from the topic entities to retrieve candidate paths under the current depth budget
$\tilde{P}_{t}(D)=\{\tilde{p} \mid  m\cdot(D-1) < \mathrm{len}(\tilde{p}) \le m\cdot D \}$,
where $m = |T(Q)|$ and $\tilde{p}=\text{Path}_{\tilde{G}_Q}(T(Q))$.
At each step, the pruning score $s_{\text{prune}}$ combines path alignment with the skyline indicator and retrieved experience, as defined under Evidence Pruning below. The top-$W_1$ paths guide the next step of local expansion.

\myparagraphunderline{Follow-up guided refine}
When local verification finds that Topic-path candidates lack sufficient support, repeating the original query tends to revisit the same neighborhood. The Hand uses the current paths to formulate a follow-up targeting the missing constraint or bridge relation.
Given $(Q, i_v)$ and the pruned candidate pool $\tilde{P}_t$, the Hand executes \textsc{GenFollowUp} locally to produce a follow-up question and indicator:
$
(Q^{+}, I^{+}) \leftarrow \textsc{GenFollowUp}(Q, i_v, \tilde{P}_t),
$
where $Q^{+}$ targets the missing information and $I^{+}$ guides the next search step.
We extract $T^{+}=T(Q^{+})$ and expand on $\tilde{G}_Q$ at greater depth (up to $D_{\max}$). New candidates enter $\tilde{P}_{r}$ for pruning and local verification against the follow-up question's constraints.

\myparagraphunderline{Prediction-driven exploration}
For cases unresolved by the first two modes, the Hand uses its parametric knowledge to propose a small set of possible answers or bridge entities and an associated indicator:
$
(\tilde{E}_{\text{pred}}, I_{\text{pred}}) \leftarrow \textsc{Predict}(Q, i_v, \tilde{P}_{t}\cup \tilde{P}_{r}),
$
and forms an augmented entity list ${T}_{\text{aug}}=T(Q)\cup \tilde{E}_{\text{pred}}$.
The Hand explores $\tilde{G}_Q$ from ${T}_{\text{aug}}$ with depth $D_{\max}$ and checks the retrieved paths against $G_Q^{\mathrm{raw}}$, retaining connected evidence that links the required entities through KG relations consistent with the current question's constraints.

\myparagraph{Evidence pruning}\label{sec:evidecePruning}
Because even an anonymized path reveals structure, \prag prunes candidates after each exploration mode through an experience-guided local filter and optional Brain-assisted selection. The filter limits the search space and the topology transmitted for remote ranking. When remote selection is unnecessary, candidates are selected locally by their pruning scores.

\myparagraphunderline{Experience-guided fuzzy selection}
Since only a small subset of generated paths is relevant, the first pruning step performs fast fuzzy selection by combining indicator alignment with retrieved experience.
For each anonymized candidate $\tilde{p}$, we compute
$
s_{\text{prune}}(\tilde{p}) = \alpha \cdot\text{DRM}(\tilde{p}, i_v)
+ (1-\alpha) \cdot\max_{\tilde{e}\in \mathcal{E}_{\text{mem}}} \text{DRM}\!\big(\tilde{p}, \tilde{e}\big),
$
where $i_v$ is the indicator at node $v$, and $\mathcal{E}_{\text{mem}}$ is the retrieved anonymized experience set. 
The top-$W_1$ paths form the filtered local candidate set.

\myparagraphunderline{Brain-assisted path selection}
Following fuzzy selection, \prag optionally calls $\mathcal{M}_{B}$ to rank and select the top-$W_{\max}$ candidates most likely to support the indicator.
$\mathcal{M}_{B}$ observes the candidate paths together with the question and indicator, all anonymized under the same session mapping. The input may contain up to $W_1$ paths, whereas $W_{\max}$ bounds the selected output. Exposure is counted over the transmitted input, including candidates excluded from the selected output.
The Hand then grounds the selected paths and checks their support against the raw KG locally (Section~\ref{sec:method:routing}).

\begin{table*}[t]
\centering
\definecolor{ctrlgBestFill}{HTML}{AFCBE8}
\definecolor{ctrlgSecondFill}{HTML}{E3EEF8}
\newcommand{\ctrlgbest}[1]{\cellcolor{ctrlgBestFill}\textbf{#1}}
\newcommand{\ctrlgsecond}[1]{\cellcolor{ctrlgSecondFill}\underline{#1}}
\caption{Accuracy (\%) on six QA benchmarks. Best: bold, darker blue. Second-best: underlined, lighter blue.}
\label{tab:main_results}
\resizebox{1\textwidth}{!}{%
\begin{tabular}{@{}lcccccccc}
\toprule
\multirow{2}{*}{\textbf{Type}} & \multicolumn{1}{c}{\multirow{2}{*}{\textbf{Method}}} & \multirow{2}{*}{\textbf{LLM}} & \multicolumn{4}{c}{\textbf{Multi-Hop KGQA}} & \textbf{Single-Hop KGQA} & \textbf{Open-Domain QA} \\ \cmidrule(lr){4-7} \cmidrule(lr){8-8} \cmidrule(lr){9-9}
 & \multicolumn{1}{c}{} &  & CWQ & WebQSP & GrailQA & QALD10-en & Simple Questions & WebQuestions \\ \toprule
\multirow{3}{*}{LLM-only} & IO prompt & GPT-3.5-Turbo & 37.6 & 63.3 & 29.4 & 42.0 & 20.0 & 48.7 \\
 & CoT \cite{wei2022cot} & GPT-3.5-Turbo & 38.8 & 62.2 & 28.1 & 42.9 & 20.3 & 48.5 \\
 & SC \cite{wang2022self} & GPT-3.5-Turbo & 45.4 & 61.1 & 29.6 & 45.3 & 18.9 & 50.3 \\ \toprule
\multirow{5}{*}{KG-centric RAG} & ToG \cite{tog1.0sun2023think} & GPT-3.5-Turbo & 58.9 & 76.2 & 68.7 & 50.2 & 53.6 & 54.5 \\
 & PoG \cite{plan-on-graph} & GPT-3.5-Turbo & 63.2 & 82.0 & 76.5 & 55.1 & 58.6 & 59.2 \\
 & GoG \cite{GoG} & GPT-3.5-Turbo & 55.7 & 78.7 & 68.2 & 50.6 & 53.1 & 54.8 \\
 & ToG-2 \cite{tog2.0ma2024think} & GPT-3.5-Turbo & 62.3 & 81.1 & 74.1 & 54.1 & 57.4 & 57.1 \\
 & ToG \cite{tog1.0sun2023think} & GPT-4 & \ctrlgbest{69.5} & 82.6 & 81.4 & 54.7 & 66.7 & 57.9 \\ \toprule
Private-KG RAG & ARoG \cite{arog_privacy} & GPT-4o-mini & 60.0 & 74.7 & 78.7 & 56.5 & 72.8 & 61.4 \\ \toprule
\multirow{6}{*}{Proposed} & \multirow{3}{*}{\makecell{\textbf{\prag} \\ Qwen3-8B [Hand]}} & GPT-4o-mini [Brain] & 61.0 & 82.4 & 80.0 & 58.6 & 77.1 & 64.9 \\
 & & DeepSeek-V3 [Brain] & 61.7 & 83.8 & 82.9 & 67.4 & 78.5 & \ctrlgsecond{71.6} \\
 & & GPT-3.5-Turbo [Brain] & 63.8 & 83.5 & 81.6 & 65.5 & \ctrlgsecond{80.9} & 67.2 \\ \cmidrule(lr){2-9}
 & \multirow{3}{*}{\makecell{\textbf{\prag} \\ Qwen3-32B [Hand]}} & GPT-4o-mini [Brain] & 62.2 & 84.3 & 83.6 & 61.0 & 79.6 & 67.0 \\
 & & DeepSeek-V3 [Brain] & 63.3 & \ctrlgsecond{85.9} & \ctrlgbest{86.1} & \ctrlgbest{70.3} & 80.2 & \ctrlgbest{74.3} \\
 & & GPT-3.5-Turbo [Brain] & \ctrlgsecond{66.2} & \ctrlgbest{86.0} & \ctrlgsecond{84.5} & \ctrlgsecond{68.8} & \ctrlgbest{83.8} & 69.0 \\ \bottomrule
\end{tabular}%
}

\end{table*}

\vspace{-3mm}
\subsection{Memory-Guided Interaction Control}\label{sec:memory}

To carry exploration experience across questions, local memory records search decisions, verified path templates, and failure notes to guide later Brain calls, search, and stopping. We describe how experience is updated, applied, and retrieved. Reuse avoids redundant search and overlapping disclosures, while each answer is grounded in current KG evidence. Remote planning receives only bounded analysis hints and path templates instantiated with current identifiers, with rare templates filtered before transmission.

Algorithm~\ref{algorithm:DualLLMController} integrates memory-gated question analysis and experience write-back with local evidence verification for each reasoning episode.

\myparagraph{Experience update}
After each reasoning episode, \prag records its outcome in $\mathcal{E}_{pool}$, including failed episodes and failed branches within successful episodes.
Records contain abstract analysis artifacts, controller decisions, and locally verified path templates together with dense embeddings:
$
\varepsilon =
\big(
\tilde{I},\,
D_{\text{predict}},\,
\tilde{\Pi},\,
\tilde{P}^{\text{tpl}},\,
\Omega,\,
e_Q,\,
e_{\tilde{I}},\,
\pi
\big),
$
where $\tilde{I}$ is the anonymized indicator, $\tilde{\Pi}$ is the mode/depth trajectory, $\tilde{P}^{\text{tpl}}$ is the set of verified anonymized path templates, and $\Omega$ summarizes sufficiency outcomes and failure notes.
Before write-back, session-local identifiers are replaced with abstract roles and coarse schema labels so that records are independent of the discarded session mapping. The local payload is encrypted at rest.
New entries are added to both $\mathcal{E}_{pool}$ and the high-frequency buffer. Redundant templates are merged, and low-value exploration traces are periodically pruned.

\myparagraphunderline{Continual adaptation}
Adaptation re-weights exemplars and periodically updates the embedding index using retrieval frequency, exposure-budget compatibility, and reasoning success. It favors experiences that aid local verification and reduces the influence of inconsistent or rarely used records on subsequent search decisions.

\myparagraph{How memory is applied}
The retrieved set $\mathcal{E}_{Q}$ is used in three places:
$i)$ Brain usage gating.
If a confident match is found, the Hand binds abstract roles to current entities and reuses the analysis and depth prior without calling the Brain.
$ii)$ Policy initialization.
The initial exploration mode and depth budget in the reasoning tree are set based on $\mathcal{E}_Q$: $
m_0,\ d_0 \leftarrow \textsc{InitPolicy}(\mathcal{E}_{Q}).
$
$iii)$ Stage transition guidance.
When verification fails, memory provides a transition suggestion based on the current controller state, i.e., whether to switch exploration mode, expand to a deeper depth, or prune the branch early: 
$
(m_{v}^{+}, d_{v}^{+}) \leftarrow \textsc{NextStep}(\mathcal{E}_{Q},\, \mathrm{state}(v)).
$
Retrieved traces steer exploration away from previously unproductive branches, while each answer is verified locally against current KG evidence.

\myparagraph{Experience retrieval}
Each active node $v$ may retrieve experience to guide Brain calls, exploration mode, and depth through a shared interface:
$
\mathcal{E}_v
=
\textsc{GetExp}(\mathcal{E}_{pool},\, Q,\, i_v,\, \pi,\, W_{\text{exp}}),
$
where $i_v$ is the node indicator, $W_{\text{exp}}$ is the retrieval width, and $\pi$ filters experiences by exposure-control mode and budget. Records include textual metadata and embeddings of questions and indicators, indexed locally with FAISS~\cite{douze2024faiss} for dual-key experience retrieval at query time.

\myparagraphunderline{Dual-key retrieval}
To match both the question semantics and the reasoning pattern represented by the indicator, we retrieve the top-$W_{\text{exp}}$ experiences from $\mathcal{E}_{pool}$ based on a hybrid score $
\textsc{Score}(\varepsilon_j)
=
\lambda_q \cdot \textsc{DRM}(Q, {q_j})
+
\lambda_I \cdot \textsc{DRM}({i_v}, {I_j}),
$
where $q_j$ and $I_j$ are the question and indicator keys of $\varepsilon_j$. Failure notes in $\Omega$ serve as warnings.
On a matching warning pattern, the controller can prune the branch early to avoid repeating a previously unsuccessful search trajectory.

\myparagraphunderline{High-frequency buffer}
To accelerate lookup and stabilize short-term behavior, a high-frequency buffer caches recently accessed exemplars, ranking them by a hybrid metric of embedding similarity and hit frequency:
$
\textsc{Score}_{\textsc{buf}}(\varepsilon_j)
=
\lambda_{\textsc{sim}}\textsc{Score}(\varepsilon_j)
+
\lambda_{\textsc{hit}}\textsc{Count}(\varepsilon_j).
$

\vspace{-3mm}
\section{Experiments}\label{sec:experiment}
\begingroup
\renewcommand{\myparagraph}[1]{\par\noindent\textbf{#1}}

We evaluate whether Ctrl-G preserves accurate KG-grounded reasoning while controlling structural and interaction exposure. Section~\ref{sec:main_results} compares answer quality across six benchmarks. We next assess structural linkability and exposure concentration (Section~\ref{sec:structural_evaluation}), followed by component effectiveness and adaptability across models and graphs (Section~\ref{sec:ablation_adaptability}). Finally, we examine memory reuse and interaction efficiency (Section~\ref{sec:memory_evaluation}). Evaluation datasets, baselines, and implementation settings are detailed in Appendix~\ref{appen:dataset_details}.

\vspace{-2.5mm}
\subsection{Main Results}\label{sec:main_results}
\vspace{-1mm}
\myparagraph{(RQ1) How effective is Ctrl-G under restricted remote access?}
To evaluate reasoning under exposure control, we compare Ctrl-G with LLM-only prompting, KG-centric RAG, and private-KG RAG on six benchmarks (Table~\ref{tab:main_results}). Across configurations, Ctrl-G achieves the highest accuracy on five datasets, outperforming the strongest baseline by up to 13.8 percentage points on QALD10-en (70.3 versus 56.5). With Qwen3-32B as Hand and GPT-3.5-Turbo as Brain, it outperforms ToG with the same remote model on all six datasets. Compared with PoG, accuracy improves from 63.2 to 66.2 on CWQ, from 82.0 to 86.0 on WebQSP, and from 76.5 to 84.5 on GrailQA. These gains support connected KG reasoning under restricted remote views.
With the same 32B Hand and GPT-4o-mini as Brain, Ctrl-G also improves over ARoG on all six benchmarks, including a 9.6-point gain on WebQSP. Replacing the Hand with Qwen3-8B lowers accuracy by 1.2--3.6 percentage points across the three Brain configurations and six datasets. Paired with GPT-4o-mini, the 8B Hand still outperforms ARoG on all six benchmarks, extending these gains to a smaller local backbone. GPT-4-based ToG retains the highest CWQ accuracy, without explicit exposure-control or privacy-preserving mechanisms. Table~\ref{tab:main_results} compares end-to-end QA performance across model configurations. RQ5 isolates the retriever within a shared architecture, while RQ6 evaluates the planner--executor configuration under matched inference budgets.

\vspace{-3.5mm}
\subsection{Structural Exposure Control}\label{sec:structural_evaluation}
\vspace{-1mm}
\myparagraph{(RQ2) Does Ctrl-G reduce structural linkability while retaining answer utility?}
\label{sec:exposure_results}\label{app:privacy_exposure}\label{app:attack_setup}
To relate answer quality to exposure, we evaluate topology-based re-identification on WebQSP and CWQ, comparing semantic masking with Ctrl-G. An honest-but-curious remote observer matches visible topic and bridge nodes against a public reference graph, without access to the session mapping. It ranks candidate identities using degree statistics and WL-style neighborhood signatures. We report ASR@1, the percentage of targets correctly identified at rank one.
Figure~\ref{fig:exposure_comparison}(a) shows that semantic masking leaves ASR@1 above 40\% on both datasets. The Ctrl-G pipeline without memory reduces it to 18.4\% on WebQSP and 19.2\% on CWQ. Within Ctrl-G, enabling memory further lowers it to 12.2\% and 12.5\%. The first comparison measures the combined retrieval and exposure controls, while the second isolates the addition of memory. Thus, masked names leave identifying structure accessible, and experience reuse further reduces its observed linkability.
We further compare complete KGQA pipelines on the same CWQ questions and backbone (Figure~\ref{fig:exposure_comparison}(b)). Ctrl-G retains accuracy close to semantic masking, at 64.2 versus 64.9, while reducing ASR@1 from 41.3\% to 12.8\%. It also achieves higher accuracy than both graph-anonymization baselines. Hay-style generalization yields the largest static anonymity set, whereas Ctrl-G has lower observed attack success. These results show that Ctrl-G retains answer utility while reducing observed linkability. Appendix~\ref{app:exploration_sensitivity} (RQ14) examines the reasoning cues retained in remote views. Relation coarsening reduces accuracy more than literal bucketization, suggesting that relation semantics help distinguish candidate reasoning paths.

\begin{figure*}[t]
\centering
\captionsetup{skip=1pt}
\begin{minipage}[t]{.49\textwidth}
\centering
\vspace{0pt}
\includegraphics[width=0.9\linewidth]{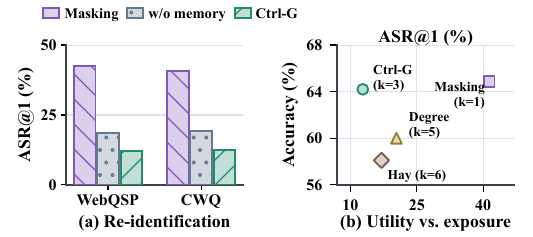}
\caption{Re-identification (a) and utility (b) in separate settings. $k$: structural anonymity set size.}
\Description{Attack success and utility-exposure operating points in separate evaluation settings. k denotes the structural anonymity set size.}
\label{fig:exposure_comparison}\label{tab:privacy_attack_asr}\label{tab:exposure_operating_points}
\end{minipage}\hfill
\begin{minipage}[t]{.49\textwidth}
\centering
\vspace{0pt}
\includegraphics[width=0.9\linewidth]{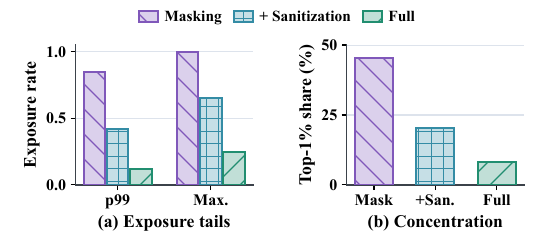}
\caption{CWQ exposure tails (a) and transmitted-node share of the most exposed 1\% of nodes (b). Lower is better.}
\Description{p99 and maximum conditional exposure rates, and the exposure share of the most frequently transmitted one percent of nodes.}
\label{fig:exposure_tails}\label{tab:layout_exposure_tail}
\end{minipage}
\end{figure*}
\myparagraph{(RQ3) Does exposure reduction extend to highly exposed nodes?}
To investigate whether lower average exposure hides risks to frequently observed nodes, we analyze the exposure distribution on CWQ. For each node in the unsanitized local working subgraph, we measure the fraction of eligible sessions in which it becomes remotely visible. All methods use the same eligible cohort, including nodes with zero exposure. We also report the share of transmitted node instances attributable to the most frequently exposed 1\% of nodes.
As shown in Figure~\ref{fig:exposure_tails}, Ctrl-G reduces p99 exposure from 0.85 to 0.12, maximum exposure from 1.00 to 0.25, and the top-1\% share from 45.2\% to 8.0\%. The reduction also reaches highly exposed nodes and limits the concentration of disclosures on a small group. Section~\ref{sec:memory_evaluation} examines linkage across repeated QA sessions.

\vspace{-1mm}
\subsection{Ablation and Adaptability Analysis}\label{sec:ablation_adaptability}
\begin{figure*}[t]
\vspace{-2mm}
\centering
\includegraphics[width=0.9\textwidth]{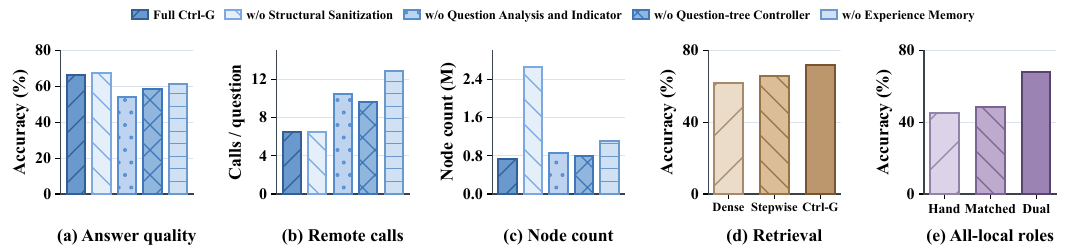}
\captionsetup{skip=1pt}
\caption{CWQ ablations (a--c), retrieval (d), and all-local roles (e) in separate settings. Panel (b) counts remote Brain calls.}
\vspace{-2mm}
\Description{Five zero-based bar panels show accuracy, remote calls, and reported node count for five component variants, then compare three retrievers and three all-local role configurations in separate controlled settings.}
\label{fig:ablation_controls}\label{tab:module_ablation_privacy}\label{fig:matched_controls}\label{tab:matched_controls}
\end{figure*}

\myparagraph{(RQ4) Which components contribute to answer quality and exposure control?}
To explain the utility and exposure results, we ablate Ctrl-G on CWQ with GPT-3.5-Turbo as Brain and utility-mode relation labels. We remove structural sanitization, question analysis and its indicator, the reasoning-tree controller, and experience memory in turn, measuring accuracy, remote calls, and node count (Figure~\ref{fig:ablation_controls}(a--c)).
The ablations separate the roles of sanitization, reasoning control, and memory. Removing structural sanitization slightly increases accuracy from 66.2 to 67.5, but raises the reported node count from 0.73M to 2.66M without reducing remote calls. Removing question analysis with its indicator or the controller instead lowers accuracy to 54.2 and 58.3, respectively, despite requiring more remote calls. Effective evidence selection and search control are both important to making productive use of remote interaction. Removing memory reduces accuracy to 61.3 and nearly doubles remote calls. The complete framework combines these functions to retain useful evidence, direct exploration, and avoid repeated search.
The five-seed indicator ablation in Appendix~\ref{app:diagnostic_uncertainty} (RQ18) uses GPT-4o-mini as Brain and yields a 10.8-point gain, with a paired bootstrap 95\% confidence interval of $[8.5,13.1]$, consistent with its role in guiding evidence retrieval.

\myparagraph{(RQ5) Does connected-path retrieval help within a fixed architecture?}
\label{app:controlled_diagnostics}
To separate retrieval quality from the surrounding architecture, we compare dense retrieval, ARoG-style stepwise retrieval, and Ctrl-G on CWQ. All three use Qwen3-32B as Hand and GPT-4-Turbo as Brain. We replace only the retriever, keeping the questions, sanitized view, controller, memory, verifier, prompts, decoding settings, and stopping rule fixed.
As shown in Figure~\ref{fig:matched_controls}(d), Ctrl-G achieves 72.0 accuracy, compared with 62.0 for dense retrieval and 65.5 for stepwise retrieval. Preserving connected paths makes inter-entity relations available for joint checking of the question's entity and relation constraints. Appendix~\ref{app:exploration_sensitivity} (RQ12--13) shows that most questions terminate in Topic-path exploration and larger budgets yield diminishing accuracy gains on CWQ.

\vspace{1mm}
\myparagraph{(RQ6) Does role decomposition help beyond additional inference budget?}
To test role specialization under matched inference use, we compare Hand-only reasoning against a local planner--executor configuration on CWQ. Both configurations use Qwen3-32B. In addition to the natural Hand-only setting, we include a control with matched retrieval access, refinement opportunities, and stopping rule. The matched Hand-only and dual-role configurations each average 10.8 total calls and 14.5K input tokens per question.
Figure~\ref{fig:matched_controls}(e) shows a 3.3-point gain from increasing Hand-only inference use, raising accuracy from 45.2 to 48.5. With the same model and inference budget, the planner--executor configuration reaches 68.0 accuracy, a 19.5-point gain over the matched Hand-only control when the local model separates planning from evidence exploration and local verification of question constraints.

\begin{figure*}[t]
\centering
\includegraphics[width=1\textwidth]{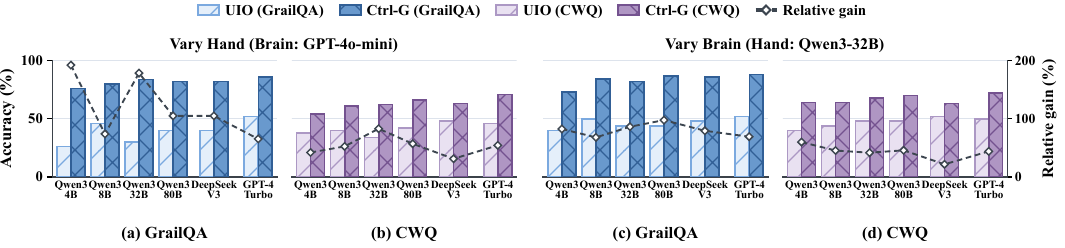}
\captionsetup{skip=1pt}
\caption{Varying Hand (a--b) or Brain (c--d). Dashed lines: relative accuracy gain over Union IO (UIO), right axis (0--200\%).}
\Description{Four panels compare UIO and Ctrl-G accuracy on GrailQA in blue and CWQ in purple. Bold model names occupy two horizontal lines. The first two panels vary Hand with GPT-4o-mini Brain, and the last two vary Brain with Qwen3-32B Hand. Dashed diamond lines show (Ctrl-G minus UIO) divided by UIO times 100, using a common zero-to-200-percent right-axis scale.}
\label{fig:backbone_main}\label{tab:backbone_main}\label{tab:ablation_backbone}
\end{figure*}

\label{app:backbone_configurations}
\vspace{1mm}

\myparagraph{(RQ7) How well does Ctrl-G adapt to different model backbones?}
To examine role adaptability, we vary the Hand and Brain backbones independently on GrailQA and CWQ, including GPT-4-Turbo in either role. Figure~\ref{fig:backbone_main}(a--b) fixes GPT-4o-mini as Brain and varies the Hand, while panels (c--d) fix Qwen3-32B as Hand and vary the Brain. We compare against Union IO (UIO), which counts a question as correct if either model answers it correctly through direct prompting.
Figure~\ref{fig:backbone_main} shows that Ctrl-G outperforms UIO across all model pairings. With Qwen3-4B as Hand, GrailQA accuracy rises from 26.0 under UIO to 76.0, showing that a small local model can support effective KG-grounded reasoning. When varying the Brain, GPT-4-Turbo achieves the highest accuracy of 88.0 on GrailQA and 72.0 on CWQ. An API-served Hand is a capability diagnostic, whereas exposure-controlled deployment keeps the Hand local (RQ8). Surpassing UIO shows that Ctrl-G answers questions missed by both models under direct prompting. RQ6 separately measures the benefit of role decomposition under a matched total inference budget per question for the same local model.

\begin{figure}[t]
\centering
\vspace{-2mm}
\captionsetup{skip=1pt}
\includegraphics[width=0.9\columnwidth]{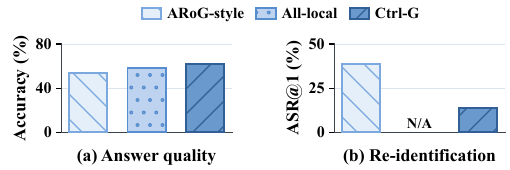}
\caption{PrimeKG accuracy and ASR. All-local exposes no remote transcript, so transcript-based ASR is inapplicable.}
\vspace{-1mm}
\label{fig:primekg_transfer}
\Description{PrimeKG answer accuracy and re-identification success compare ARoG-style, all-local, and Ctrl-G pipelines. All-local has no remote transcript.}
\end{figure}

\myparagraph{(RQ8) What does a remote Brain add to all-local reasoning?}
\label{app:all_local_deployment}
To assess whether remote assistance is necessary, we fix Qwen3-32B as Hand and compare a local Qwen3-32B Brain with a remote GPT-4-Turbo Brain under the same exploration and verification pipeline. In Figure~\ref{fig:backbone_main}(c--d), all-local Ctrl-G reaches 82.0 accuracy on GrailQA and 68.0 on CWQ with zero remote calls. Replacing only the Brain with GPT-4-Turbo raises accuracy to 88.0 and 72.0. Together with RQ6, this separates role specialization from the additional 6.0- and 4.0-point gains of a stronger Brain. Ctrl-G can either operate entirely locally or use remote planning while retaining local grounding and verification against the current KG evidence for each question.

\myparagraph{(RQ9) Does Ctrl-G transfer beyond Freebase?}
\label{app:cross_domain}
To test transfer beyond Freebase, we evaluate Ctrl-G on the public biomedical PrimeKG~\cite{chandak2023primekg} (Figure~\ref{fig:primekg_transfer}). Ctrl-G improves accuracy over the ARoG-style pipeline by 8.4 points while reducing ASR@1 from 38.9\% to 13.6\%. It also exceeds all-local reasoning by 3.9 points. All-local ASR is inapplicable because no transcript reaches a remote model. These results extend the utility and exposure benefits to biomedical QA. Ctrl-G retains local grounding and remote-view control on a graph with different entity types, relation semantics, and schema structure, using retrieved biomedical evidence to verify candidate answers against the question's entity and relation constraints.

\vspace{-2mm}
\subsection{Memory-Guided Interaction Control}\label{sec:memory_evaluation}
\vspace{-1mm}

\myparagraph{(RQ10) Does memory reduce repeated disclosure?}
\label{sec:interaction_diagnostic}\label{app:interaction_exposure}
We assess memory reuse across complete reasoning runs through answer quality, remote calls, and topology-linkage success. Enabling memory raises accuracy from 61.3 to 66.2 and reduces remote calls from 12.8 to 6.5 (Figure~\ref{fig:ablation_controls}(a--b)), alongside lower ASR in Figure~\ref{fig:exposure_comparison}(a). Appendix~\ref{app:memory_behavior} (RQ15--16) reports diminishing cold-start gaps as experience accumulates and faster lookup with the high-frequency buffer. Its audit of 12,480 memory records (RQ17) finds no raw identifiers or session mappings in stored traces.
We further compare the complete pipelines in a ten-session linkage stress test with fresh identifiers. ASR@1 decreases from 62.1\% under semantic masking to 20.2\% with full Ctrl-G. This joint pipeline comparison extends the observed benefit to repeated interactions, where fresh identifiers still leave recurring structure available for linkage. These results support controlling the graph content of each transmitted view together with the accumulation of evidence across successive remote calls.
\begin{figure}[t]
\centering
\vspace{-2mm}
\captionsetup{skip=1pt,justification=raggedright,singlelinecheck=false}
\includegraphics[width=0.9\columnwidth]{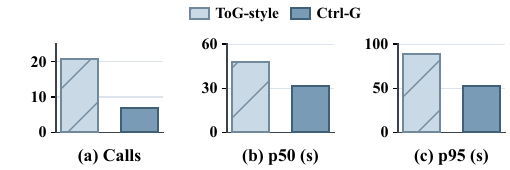}
\caption{CWQ remote calls and end-to-end latency. p50 and p95 denote the median and 95th percentile latencies.}
\vspace{-1mm}
\Description{Remote calls, median latency, and 95th-percentile latency for ToG-style traversal and Ctrl-G.}
\label{fig:interaction_cost}\label{tab:layout_latency}
\end{figure}
\myparagraph{(RQ11) Does Ctrl-G reduce remote calls and latency?}
To evaluate the practical cost of remote-assisted reasoning, we measure API calls and end-to-end latency on CWQ. In the primary evaluation, Ctrl-G requires 6.5 remote calls on average, compared with 12.8 without memory and 21.7 for ToG. We further compare Ctrl-G with ToG-style traversal in a separate latency evaluation (Figure~\ref{fig:interaction_cost}). Remote calls decrease from 20.9 to 6.8, while median latency falls from 47.8 to 31.6 seconds and p95 latency from 88.4 to 52.4 seconds.
Median and p95 latency fall by about 34\% and 41\%, respectively. These reductions benefit both typical queries and those in the slower tail. Appendix~\ref{app:transfer_deployment} (RQ19) reports remote tokens and total calls. Ctrl-G uses fewer remote tokens than all four baselines on GrailQA. Local Hand tokens are reported separately in the same appendix.

To further evaluate Ctrl-G, Appendix~\ref{appendix:exp} examines exploration, remote views, memory, reliability, and computational cost.

\endgroup
\section{Conclusion}
\label{sec:conclusion}
We present Ctrl-G, a training-free KG-RAG framework for controlling structural and interaction exposure. It combines structural sanitization, connected-path exploration, and experience reuse, with evidence grounded and verified locally against the original KG. Experiments on six QA benchmarks show strong answer quality across model configurations. Further analyses show reduced structural linkability and remote-call demand. Ctrl-G thus controls what the remote model observes and when its assistance is needed, supporting accurate multi-hop question answering over private KGs.

\clearpage
\appendix
\begingroup
\SetCommentSty{textbf}
\SetArgSty{textup}
\SetFillComment
\printsemicolon
\section{Algorithms}\label{appendix:alg}
\subsection{Hierarchical Reasoning}\label{appendix:alg:routing}
Algorithm~\ref{algorithm:DualLLMController} details dual-LLM hierarchical reasoning and question analysis (Section~\ref{sec:method:routing}).
\textsc{RemoteHints} instantiates filtered analysis hints and path templates using current identifiers. \textsc{CanContinue} tests whether further exploration remains within the search limits.
\vspace{2mm}

\begin{algorithm}[H]
{
{
\SetVline
\small
\caption{{\small{DualLLMHierarchicalReasoning}}}\label{algorithm:DualLLMController}

\Input{Question $Q$, raw subgraph $G^{\mathrm{raw}}_Q$, anonymized view $\tilde{G}_Q$, topic entities $T(Q)$, session mapping $\phi_Q$, exposure-control configuration $\pi$, experience pool $\mathcal{E}_{pool}$, limits $D_{\max}, W_1, W_{\max}, W_{\exp}$}
\Output{Answer $A$, evidence $E_{\text{final}}$, sufficiency status, updated pool $\mathcal{E}_{pool}$}

\tcc{Memory-gated Brain usage}
\State{$\mathcal{E}_{Q} \leftarrow \text{GetExp}(\mathcal{E}_{pool}, Q, I_{\text{root}}, \pi, W_{\exp})$}
\State{$\text{CallBrain} \leftarrow \text{GateBrainUsage}(\mathcal{E}_{Q})$}

\State{$(\mathcal{T}_Q, \mathcal{I}_Q, D_{\text{predict}}) \leftarrow \text{QuestionAnalysis}(Q, T(Q), \tilde{G}_Q, \phi_Q, \mathcal{E}_{Q}, \text{CallBrain})$}

\tcc{Initialize reasoning}
\State{$(m_0,d_0) \leftarrow \text{InitPolicy}(\mathcal{E}_{Q})$}
\State{$\text{InitTree}(\mathcal{T}_Q, m_0, d_0, D_{\text{predict}})$}

\Repeat{$\text{sufficient} \lor \neg\text{CanContinue}(\mathcal{T}_Q)$}{
\ForEach{$v \in \text{TraverseTopDown}(\mathcal{T}_Q)$}{
  \State{\textbf{if} $\text{status}(v)=\text{Pruned}$ \textbf{then continue}}
  \State{$(\text{status}(v), E_v) \leftarrow \text{NodeVerificationLoop}(v, Q, G^{\mathrm{raw}}_Q, \tilde{G}_Q$, $\phi_Q, \mathcal{I}_Q, \pi, \mathcal{E}_{pool}, D_{\max}, W_1, W_{\max}, W_{\exp})$}
  \State{$\text{AttachEvidence}(\mathcal{T}_Q, v, E_v)$}
}

\State{$A, E_{\text{final}}, \text{sufficient} \leftarrow \text{SynthesizeAndCheck}(Q, \mathcal{T}_Q)$}
\lIf{$\neg\text{sufficient} \land \text{CanContinue}(\mathcal{T}_Q)$}{$\text{UpdateTree}(\mathcal{T}_Q)$\;}
}
  \tcc{Record verified templates and all episode outcomes}
  \State{$\tilde{\Pi} \leftarrow \text{ExtractTrajectory}(\mathcal{T}_Q)$}
  \State{$\tilde{P}^{\text{tpl}} \leftarrow \text{ExtractVerifiedTemplates}(\mathcal{T}_Q)$}
  \State{$\Omega \leftarrow \text{SummarizeOutcomes}(\mathcal{T}_Q)$}
  \State{$\mathcal{E}_{pool} \leftarrow \text{ExperienceUpdate}(\mathcal{E}_{pool}, Q, \mathcal{I}_Q$, $D_{\text{predict}}, \tilde{\Pi}, \tilde{P}^{\text{tpl}}, \Omega, \pi)$}
\State{\textbf{return} $A, E_{\text{final}}, \text{sufficient}, \mathcal{E}_{pool}$}

\vspace{1mm}
\textbf{Procedure} \text{QuestionAnalysis}$(Q, T(Q)$, $\tilde{G}_Q$, $\phi_Q$, $\mathcal{E}_{Q}$, $\text{CallBrain})$\\
\SetVline
\small
\If{$\text{HasReusableAnalysis}(\mathcal{E}_{Q})$}{
  \State{$(\mathcal{T}_Q,\mathcal{I}_Q,D_{\text{predict}}) \leftarrow \text{ReuseAnalysis}(\mathcal{E}_{Q},T(Q),\phi_Q)$}
}
\Else{
  \If{$\text{CallBrain}$}{
    \State{$\tilde{Q} \leftarrow \text{AnonText}(Q,\phi_Q)$}
    \State{$\tilde{T}(Q) \leftarrow \phi_Q(T(Q))$}
    \State{$\tilde{H}_Q \leftarrow \text{RemoteHints}(\mathcal{E}_{Q},\phi_Q)$}
    \State{$(\tilde{\mathcal{T}}_Q,\tilde{\mathcal{I}}_Q,D_{\text{predict}}) \leftarrow \mathcal{M}_B(\tilde{Q}, \tilde{T}(Q), \tilde{H}_Q)$}
    \State{$(\mathcal{T}_Q, \mathcal{I}_Q) \leftarrow \text{ApplyMap}((\tilde{\mathcal{T}}_Q, \tilde{\mathcal{I}}_Q), \phi_Q^{-1})$}
  }
  \Else{
    \State{$(\mathcal{T}_Q,\mathcal{I}_Q,D_{\text{predict}}) \leftarrow \mathcal{M}_H(Q, T(Q))$}
    \State{$(\tilde{\mathcal{T}}_Q,\tilde{\mathcal{I}}_Q) \leftarrow \text{ApplyMap}((\mathcal{T}_Q,\mathcal{I}_Q),\phi_Q)$}
  }
}
\State{\textbf{return} $\mathcal{T}_Q,\mathcal{I}_Q,D_{\text{predict}}$}
}
}
\end{algorithm}

\newpage
\subsection{Evidence Exploration and Pruning}\label{appendix:alg:exploration}
We present the detailed procedures for evidence exploration and path pruning (Section~\ref{sec:Method:retrieval_pruning}) in Algorithms~\ref{algorithm:ExploreAndPrune} and~\ref{algorithm:EvidencePruning}, respectively.
The Hand executes both \textsc{GenFollowUp} and \textsc{Predict} locally.
Pools start empty and persist across controller steps at each node.
The procedures share the local session mapping $\phi_Q$ during retrieval.
\vspace{2mm}

\begin{algorithm}[H]
{
{
\SetVline
\small
\caption{{\small{EvidenceExploration}}}\label{algorithm:ExploreAndPrune}

\Input{Anonymized KG $\tilde{G}_Q$, question $Q$, topics $T(Q)$, indicator $i_v$, predicted depth $D_{\text{predict}}$, exposure-control configuration $\pi$, action $(m,d)$, candidate pools $(\tilde{P}_t,\tilde{P}_r,\tilde{P}_p)$, experience pool $\mathcal{E}_{pool}$, limits $D_{\max}, W_1, W_{\max}, W_{\exp}$}
\Output{Pruned candidates $\tilde{P}_v$, updated pools $(\tilde{P}_t,\tilde{P}_r,\tilde{P}_p)$}

\tcc{Topic-path exploration with depth expansion}
\If{$m=\text{Topic} $}{
  \For{$D \leftarrow \min(D_{\text{predict}}, D_{\max})$ \KwTo $d$}{
    \State{$\tilde{P} \leftarrow \text{TreeBiBFS}(\tilde{G}_Q, T(Q), D,W_{\max},Q,i_v)$}

    \State{$\tilde{P}_t \leftarrow \tilde{P}_t \cup \tilde{P}$}
  }
  \State{$\tilde{P}_t \leftarrow \text{EvidencePruning}(Q, i_v, \tilde{P}_t, \pi, \mathcal{E}_{pool}, W_1, W_{\max}, W_{\exp})$}
}
\tcc{Follow-up guided refinement}
\If{$m=\text{Refine}$}{
  \State{$(Q^{+}, I^{+}) \leftarrow \text{GenFollowUp}(Q, i_v, \tilde{P}_t)$}
  \State{$T^{+} \leftarrow \text{TopicEntityRecognize}(Q^{+})$}
  \State{$\tilde{P}_r \leftarrow \text{TreeBiBFS}(\tilde{G}_Q, T^{+}, D_{\max},W_{\max},Q^+,I^+)$}
  \State{$\tilde{P}_r \leftarrow\text{EvidencePruning}(Q^{+}, I^{+}, \tilde{P}_r, \pi, \mathcal{E}_{pool}, W_1, W_{\max}, W_{\exp})$}
}

\tcc{Prediction-driven exploration}
\If{$m=\text{Predict}$}{
  \State{$\tilde{P}_{\text{in}} \leftarrow \tilde{P}_t \cup \tilde{P}_r$}
  \State{$(\tilde{E}_{\text{pred}}, I_{\text{pred}}) \leftarrow \text{Predict}(Q, i_v, \tilde{P}_{\text{in}})$}
  \State{$T_{\text{aug}} \leftarrow T(Q) \cup \tilde{E}_{\text{pred}}$}
  \State{$\tilde{P}_p \leftarrow \text{TreeBiBFS}(\tilde{G}_Q, T_{\text{aug}}, D_{\max}, W_{\max},Q,I_{\text{pred}})$}
  \State{$ \tilde{P}_p \leftarrow\text{EvidencePruning}(Q, I_{\text{pred}}, \tilde{P}_p, \pi, \mathcal{E}_{pool}, W_1, W_{\max}, W_{\exp})$}
}

\State{$\tilde{P}_v \leftarrow \tilde{P}_t \cup \tilde{P}_r\cup \tilde{P}_p$}
\State{\textbf{return} $\tilde{P}_v, (\tilde{P}_t,\tilde{P}_r,\tilde{P}_p)$}
}
}
\end{algorithm}

\begin{algorithm}[H]
{
{
\SetVline
\small
\caption{{\small{EvidencePruning}}}\label{algorithm:EvidencePruning}

\Input{Question $Q$, indicator $I$, candidates $\tilde{P}$, exposure-control configuration $\pi$, experience pool $\mathcal{E}_{pool}$, widths $W_1, W_{\max}, W_{\exp}$}
\Output{Pruned candidates $\tilde{P}_{\text{out}}$}
\tcc{Experience-guided local pruning}
\State{$\mathcal{E}_{\text{mem}} \leftarrow \text{GetExp}(\mathcal{E}_{pool}, Q, I, \pi, W_{\exp})$}
\ForEach{$\tilde{p}\in \tilde{P}$}{
  \State{$s_{\text{prune}}(\tilde{p}) \leftarrow
  \alpha \cdot \text{DRM}(\tilde{p}, I)
  + (1-\alpha)\cdot \max_{\tilde{e}\in \mathcal{E}_{\text{mem}}}\text{DRM}(\tilde{p}, \tilde{e})$}
}
\State{$\tilde{P}^{\star} \leftarrow \text{TopK}(\tilde{P}, s_{\text{prune}}, W_1)$}

\tcc{Optional Brain-assisted selection}
\If{$\text{NeedBrainSelect}(\pi,\tilde{P}^{\star})$}{
  \State{$(\tilde{Q},\tilde{I}) \leftarrow \text{AnonText}((Q,I),\phi_Q)$}
  \State{$\tilde{P}_{\text{out}} \leftarrow \mathcal{M}_B(\tilde{Q}, \tilde{I}, \tilde{P}^{\star}, W_{\max})$}
}
\lElse{$\tilde{P}_{\text{out}} \leftarrow \text{TopK}(\tilde{P}^{\star}, s_{\text{prune}}, W_{\max})$\;}

\State{\textbf{return} $\tilde{P}_{\text{out}}$}
}
}
\end{algorithm}
\clearpage
\endgroup

\begingroup
\makeatletter
\let\ctrlg@startsection\@startsection
\def\@startsection#1#2#3#4#5#6{%
  \ctrlg@startsection{#1}{#2}{#3}{-6pt}{3pt}{#6}}
\makeatother
\setlength{\textfloatsep}{4pt plus 1pt minus 1pt}
\setlength{\dbltextfloatsep}{4pt plus 1pt minus 1pt}
\setlength{\floatsep}{3pt plus 1pt minus 1pt}
\setlength{\dblfloatsep}{3pt plus 1pt minus 1pt}
\setlength{\intextsep}{3pt plus 1pt minus 1pt}
\section{Additional Experiments}\label{appendix:exp}
\renewcommand{\myparagraph}[1]{\par\noindent\textbf{#1}}
\subsection{Exploration and View Sensitivity}\label{app:exploration_sensitivity}
\begin{figure*}[t]
\centering
\captionsetup{skip=1pt,justification=raggedright,singlelinecheck=false}
\begin{minipage}[t]{.3332\textwidth}
\centering
\includegraphics[width=\linewidth]{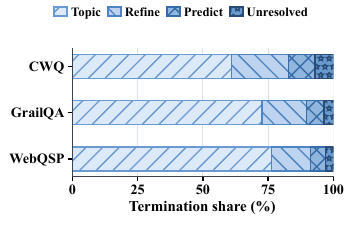}
\caption{Termination shares by mode, including unresolved questions.}
\Description{Stacked horizontal bars show Topic-path, Follow-up guided refine, Prediction-driven, and unresolved termination shares for CWQ, GrailQA, and WebQSP. Most questions terminate during Topic-path.}
\label{fig:phase_usage}
\end{minipage}\hfill
\begin{minipage}[t]{.3038\textwidth}
\centering
\includegraphics[width=\linewidth]{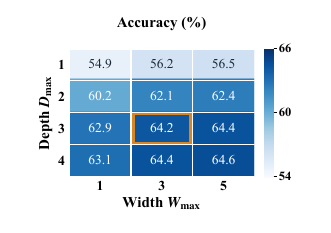}
\caption{Search bounds on CWQ. Orange box: default setting.}
\Description{Accuracy for twelve depth and width combinations. The orange outline marks the default depth and width of three.}
\label{fig:layout_diagnostic_panels}
\end{minipage}\hfill
\begin{minipage}[t]{.343\textwidth}
\centering
\includegraphics[width=\linewidth]{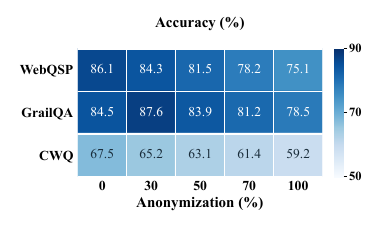}
\caption{Answer accuracy across three datasets at five anonymization ratios.}
\Description{Accuracy at five anonymization ratios for WebQSP, GrailQA, and CWQ.}
\label{fig:anonymization_impact}
\end{minipage}
\end{figure*}
\myparagraph{(RQ12) Exploration phase usage.}
\label{app:phase_usage}
Figure~\ref{fig:phase_usage} reports question termination modes on CWQ, GrailQA, and WebQSP. Topic-path accounts for 61.0\%--76.3\% of questions. Successively enabling Follow-up guided refine and Prediction-driven exploration increases average calls from 3.9 to 5.7 to 7.3 on CWQ, from 3.2 to 4.4 to 5.5 on GrailQA, and from 3.0 to 4.2 to 5.2 on WebQSP before memory gating. Memory reduces the complete CWQ configuration to 6.8 calls. Follow-up generation and prediction run locally on the Hand, while the resulting candidates may require further Brain-assisted ranking. Staged exploration reserves further interaction for questions that require additional evidence beyond the initial Topic-path search to resolve missing constraints or bridge relations.

\vspace{1mm}

\myparagraph{(RQ13) Sensitivity to exploration budgets.}
\label{app:phase_diagnostics}
To assess the trade-off between search capacity and exposure, we vary $D_{\max}$ and $W_{\max}$ on CWQ (Figure~\ref{fig:layout_diagnostic_panels}). At $W_{\max}=3$, increasing depth from 1 to 3 raises accuracy from 56.2 to 64.2. A further increase to 4 adds only 0.2 points, but raises calls from 6.8 to 8.4 and unique visible nodes per question from 142 to 184. Increasing width to 5 also adds only 0.2 points, with 8.1 calls and 181 visible nodes. Varying the candidate width $W_1$ from 40 to 80 to 120 gives accuracies of 63.6, 64.2, and 64.3, with 6.5, 6.8, and 7.1 calls. The default lies near an accuracy plateau: broader search exposes substantially more structure for little additional answer utility while requiring more remote calls to process and rank the newly expanded candidate evidence.

\begin{figure}[t]
\centering
\captionsetup{skip=1pt}
\includegraphics[width=\columnwidth]{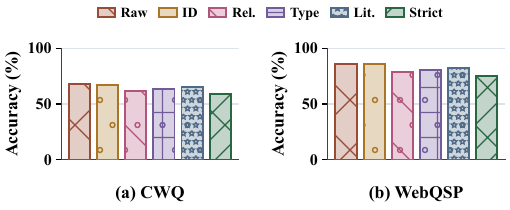}
\caption{Remote-view ablations on CWQ (a) and WebQSP (b). Strict combines semantic and structural sanitization. ID, Rel., Type, and Lit. denote identifier replacement, relation clustering, entity types, and literal bucketization.}
\Description{Six remote-view variants on each dataset: raw view, identifier replacement, relation clustering, entity types, literal bucketization, and the strict combined configuration.}
\label{fig:appendix_six}\label{fig:signal_ablation}\label{tab:signal_ablation}
\end{figure}
\vspace{1mm}

\myparagraph{(RQ14) Remote-view signal ablations.}
\label{app:signal_ablation}\label{exp:privacy_robustness}
To identify which signals support remote reasoning, we vary the anonymization ratio and ablate the content retained in remote views. Increasing anonymization from 0\% to 100\% reduces CWQ accuracy from 67.5\% to 59.2\%, while WebQSP retains 75.1\% under full anonymization (Figure~\ref{fig:anonymization_impact}). GrailQA peaks at 87.6\% with 30\% anonymization before decreasing to 78.5\%, showing a non-monotonic relationship between anonymization and accuracy. In Figure~\ref{fig:signal_ablation}(a--b), replacing entity identifiers while retaining labels keeps accuracy close to the plaintext view. Relation clustering causes a larger decrease than literal bucketization, suggesting that relation semantics help distinguish candidate paths. Combining entity anonymization, relation clustering, literal bucketization, and structural sanitization retains 59.2 accuracy on CWQ and 75.1 on WebQSP. The trade-off reflects both the amount of content masked and the reasoning cues preserved in the remote view for guiding evidence selection.

\subsection{Memory Behavior and Audit}\label{app:memory_behavior}
\begin{figure*}[t]
\centering
\captionsetup{skip=1pt}
\begin{minipage}[t]{.8019\textwidth}
\centering
\includegraphics[width=\linewidth]{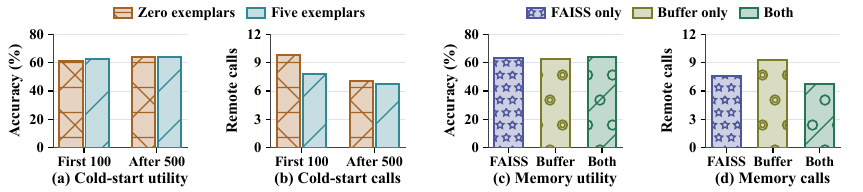}
\caption{Memory initialization (a--b) and component ablations (c--d) on CWQ. Checkpoints count processed questions. Calls count remote Brain invocations.}
\Description{Four panels compare accuracy and remote calls for memory initialization and components. Checkpoints count questions.}
\label{fig:memory_panels}\label{tab:layout_cold_start}
\label{fig:memory_components}\label{tab:memory_components}
\end{minipage}\hfill
\begin{minipage}[t]{.1881\textwidth}
\centering
\includegraphics[width=\linewidth]{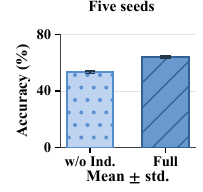}
\captionsetup{justification=raggedright,singlelinecheck=false}
\caption{CWQ \mbox{indicator} ablation.}
\Description{CWQ indicator-ablation accuracy over five seeds. Whiskers show standard deviation across five seeds.}
\label{fig:indicator_reliability}
\end{minipage}
\end{figure*}

\vspace{1mm}

\myparagraph{(RQ15) Memory initialization.}
To evaluate dependence on initial experience, we compare zero and five exemplars along a CWQ question stream (Figure~\ref{fig:memory_panels}(a--b)). During the first 100 questions, five exemplars improve accuracy from 61.0 to 63.0 and reduce calls from 9.8 to 7.8. After 500 questions, the gaps narrow to 0.3 accuracy points and 0.3 remote calls. Initial exemplars mainly ease the cold start. As verified traces accumulate, an initially empty memory approaches the initialized configuration in answer accuracy and remote-call demand over the observed question stream.

\vspace{1mm}

\myparagraph{(RQ16) Memory component ablations.}
To separate the roles of the FAISS index and high-frequency buffer, we ablate each on CWQ (Figure~\ref{fig:memory_components}(c--d)). FAISS-only obtains higher accuracy and fewer remote calls than buffer-only. Adding the buffer to FAISS changes accuracy from 63.7 to 64.2, reduces calls from 7.6 to 6.8, and cuts local lookup latency from 6.8 to 2.4\,ms. The clearest benefit is cheaper access to frequently reused experience and lower remote-call demand, alongside a modest improvement in grounded answer accuracy within the same memory-based reasoning pipeline.

\myparagraph{(RQ17) Persistent-memory audit.}
To examine the information retained for reuse, we audit all 12,480 memory records. None contains raw identifiers, session mappings, unbucketized sensitive literals, or prohibited relation labels. We filter 73 rare templates (0.58\%) before remote-facing use because abstract patterns can still be distinctive. Memory, its index, and access statistics remain local. Retrieved patterns are remapped under fresh identifiers and re-verified before reuse. Stored exploration choices and failure patterns can guide search modes and depths or help avoid failed branches. The audit complements RQ10's interaction analysis by examining these reusable decisions, while each answer remains grounded in current KG evidence rather than entity bindings from an earlier question.

\vspace{-2mm}
\subsection{Reliability and Computational Cost}\label{app:transfer_deployment}
\vspace{-1mm}

\myparagraph{(RQ18) Is the indicator gain stable across repeated runs?}
\label{app:diagnostic_uncertainty}
To examine the stability of the indicator's contribution, we repeat its ablation on CWQ across five seeds (Figure~\ref{fig:indicator_reliability}). Full Ctrl-G achieves $64.2\pm0.7$ accuracy, compared with $53.4\pm0.9$ without the indicator, where the whiskers denote standard deviation across seeds. The 10.8-point improvement has a question-paired bootstrap 95\% confidence interval of $[8.5,13.1]$. Resampling groups all seed outcomes by question to preserve within-question dependence. The interval characterizes uncertainty conditional on the recorded runs and their memory-update orders. Its positive range supports the observed indicator gain in CWQ runs under this evaluation protocol.

\begin{figure*}[t]
\centering
\captionsetup{skip=1pt}
\includegraphics[width=\textwidth]{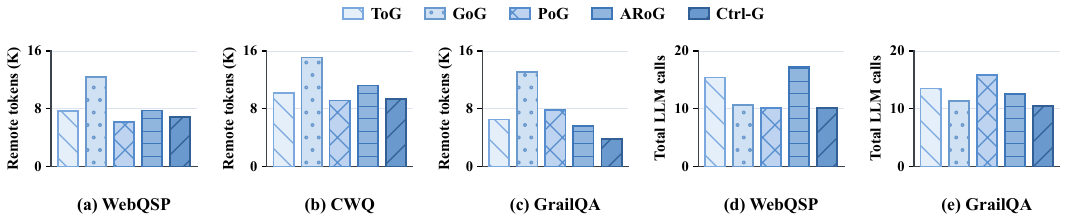}
\caption{Remote token usage (a--c, Brain only for Ctrl-G) and total LLM calls (d--e, Hand plus Brain for Ctrl-G) per question.}
\Description{Five panels compare ToG, GoG, PoG, ARoG, and Ctrl-G. The first three show remote input-plus-output tokens on WebQSP, CWQ, and GrailQA, counting only the Brain for Ctrl-G. The last two show total LLM calls on WebQSP and GrailQA, including both Ctrl-G roles in one unstacked bar.}
\label{fig:role_tokens}\label{fig:cost_comparison}
\end{figure*}

\myparagraph{(RQ19) What is the remote inference demand?}
\label{app:role_tokens}
Figure~\ref{fig:cost_comparison} compares remote input-plus-output tokens and total LLM calls per question. On CWQ, the Brain uses 9.3K tokens, compared with 10.2K for ToG, 15.1K for GoG, 9.2K for PoG, and 11.2K for ARoG. On GrailQA, it uses 3.8K tokens, below the 5.6K--13.1K reported for these baselines. This comparison complements the remote-call reductions in RQ11. The call panels count both Hand and Brain for Ctrl-G, totaling 10.2 calls on WebQSP, close to PoG's 10.1 and below ToG's 15.4. Its 10.5 calls on GrailQA are fewer than all four baselines. The local Hand additionally processes 10.5K, 11.3K, and 8.2K tokens per question on WebQSP, CWQ, and GrailQA, respectively. On CWQ, the two roles together use 20.6K tokens per question. The counts distinguish local graph exploration and evidence verification from remote planning and candidate ranking, allowing the inference demand of each role to be examined separately.

\label{app:cost_details}
\vspace{-2mm}
\section{Experiment Details}\label{appen:dataset_details}
\vspace{-1mm}

\myparagraph{Datasets and evaluation.}
CWQ and GrailQA test complex and compositional QA. WebQSP and WebQuestions use Freebase, QALD10-en covers English QA over linked data, and SimpleQuestions tests single-relation QA. Table~\ref{tab:appendix_dataset} lists references, sizes, and licenses. Topic entities are linked locally. We report Hits@1 for QA and attack success for re-identification. Ablations use the same 500-question subsets and inference conditions across variants.

\myparagraph{Baselines.}
Table~\ref{tab:main_results} compares LLM-only, KG-centric, and private-KG methods. ARoG combines entity anonymization, relation-centric abstraction, and structured concept paths for remote QA.

\myparagraph{Implementation settings.}
Unless otherwise specified, experiments use Qwen3-32B as the local Hand and GPT-4o-mini as the remote Brain. Table~\ref{tab:main_results} lists the main backbone combinations. Additional backbones are Qwen3-4B, Qwen3-Next-80B-A3B-Instruct (abbreviated Qwen3-80B), and GPT-4-Turbo. Exploration and final-generation temperatures are 0.4 and 0, with at most 256 output tokens. We set $D_{\max}=W_{\max}=3$, $W_1=80$, and semantic weight $\lambda_{\rm sem}=0.6$. Memory retrieval uses $\lambda_q=\lambda_I=0.5$, $\lambda_{\rm sim}=0.7$, $\lambda_{\rm hit}=0.3$, and $W_{\exp}=5$. Memory starts with five exemplars and maintains a 1,000-entry high-frequency buffer for local lookup.

\myparagraph{Sanitization and memory.}
Structural clusters use coarse types, bucketized degrees and literals, and two-hop WL-style signatures. Unless otherwise specified, experiments use utility-oriented mode, which retains relation labels. The stricter mode coarsens them. Structural pruning combines indicator relevance, path participation, and distance to topic anchors. The Hand expands supernodes and verifies edges locally. Memory records replace identifiers with abstract roles and coarse labels, remain encrypted locally, and are re-verified against the KG after remapping to fresh identifiers.

\par\smallskip
\par\endgroup

\begin{table}[t]
\centering\scriptsize
\vspace{-2mm}
\caption{Evaluation datasets.}
\label{tab:appendix_dataset}
\setlength{\tabcolsep}{2pt}
\resizebox{\columnwidth}{!}{%
\begin{tabular}{@{}lcccccc@{}}
\toprule
Dataset & \makecell{CWQ\\\cite{talmor-berant-2018-web}} & \makecell{WebQSP\\\cite{yih-etal-2016-value}} & \makecell{GrailQA\\\cite{grailqa}} & \makecell{QALD10-en\\\cite{usbeck2024qald}} & \makecell{Simple\\Questions~\cite{bordes2015simplequestions}} & \makecell{Web\\Questions~\cite{berant-etal-2013-semantic}}\\\midrule
Test & 1,000 & 1,639 & 1,000 & 333 & 1,000 & 2,032\\
Train & 27,734 & 3,098 & 44,337 & -- & 14,894 & 3,778\\
License & Apache-2.0 & MSR-LA & MIT & MIT & CC BY 3.0 & CC BY 4.0\\\bottomrule
\vspace{-2.4mm}

\end{tabular}%
\vspace{-2mm}
}
\end{table}

\end{document}